\documentclass[journal]{IEEEtran}
\usepackage{graphics} 
\usepackage{subfigure}
\usepackage{epsfig} 
\usepackage{multicol}
\usepackage{amssymb}  
\usepackage{algorithm}
\usepackage{algorithmicx}
\usepackage{algpseudocode}
\usepackage{epstopdf}
\usepackage{multirow}
\newtheorem{theorem}{Theorem}

\usepackage{cite}

\ifCLASSINFOpdf
\else
\fi
\usepackage{amsmath}
\usepackage{array}
\begin{document}
	%
	\title{System Design and Control of an Apple Harvesting Robot}
	%
	%
	%
	
	\author{Kaixiang Zhang$^{*}$,
		Kyle Lammers$^{*}$,
		Pengyu Chu,
		Zhaojian Li,
		and Renfu Lu 
		\thanks{Kaixiang Zhang, Kyle Lammers, Pengyu Chu, and Zhaojian Li are with the Department of Mechanical Engineering, Michigan State University, East Lansing, MI 48824, USA (e-mail: zhangk64@msu.edu; lammer18@msu.edu; chupengy@msu.edu; lizhaoj1@egr.msu.edu).}
		\thanks{Renfu Lu is with  the United States Department of Agriculture Agricultural Research Service, East Lansing, MI 48824, USA (e-mail: renfu.lu@usda.gov).}
		\thanks{* Both authors contributed equally to this work.}
	}
	
	%
	%

	\markboth{}%
	{Shell \MakeLowercase{\textit{et al.}}: Bare Demo of IEEEtran.cls for IEEE Journals}
	%



	\maketitle
	
	\begin{abstract}
		There is a growing need for robotic apple harvesting due to decreasing availability and rising cost in labor. 
		Towards the goal of developing a viable robotic system for apple harvesting, this paper presents synergistic mechatronic design and motion control of a robotic apple harvesting prototype, which lays a critical foundation for future advancements. 
		Specifically, we develop a deep learning-based fruit detection and localization system using an RGB-D camera. 
		A three degree-of-freedom manipulator is then designed with a hybrid pneumatic/motor actuation mechanism to achieve fast and dexterous movements. A vacuum-based end-effector is used for apple detaching.
		These three components are integrated into a robotic apple harvesting prototype with simplicity, compactness, and robustness. Moreover, a nonlinear velocity-based control scheme is developed for the manipulator to achieve accurate and agile motion control.   
		Test experiments are conducted to demonstrate the performance of the developed apple harvesting robot.
	\end{abstract}
	
	
	%
	\IEEEpeerreviewmaketitle

	\section{Introduction}
	%
	%
	%
	%
	\IEEEPARstart{A}{pple} harvesting is a physically strenuous and labor intensive task. It is estimated that the seasonal agricultural workforce required for apple harvesting in the U.S. is more than 10 million worker hours annually, accounting for about 15\% of the total production cost \cite{gallardo2012}.
	The long-term proﬁtability and sustainability of the apple industry has been eroded due to the decreasing availability and rising cost in the labor market. Moreover, manual picking activities expose the workers to great risks of ergonomic injury and musculoskeletal pain, as manual picking involves extensive repetitive body motions and awkward postures (especially when picking fruits at high locations or deep in the canopy, and repetitively ascending and descending on ladders with heavy loads) \cite{fathallah2010}. As such, there is an imperative need for automated apple picking to address the aforementioned concerns. 
	
	Existing automated apple harvesting systems can be generally categorized as shake-and-catch harvesting \cite{peterson1999,de2015,he2017,he2019} or fruit-by-fruit harvesting \cite{baeten2008,de2011,Davidson2016IROS,silwal2017}.
	A shake-and-catch harvesting system vibrates the branches or trunk of the tree to detach apples, and a catching device is then employed to catch the falling apples. 
	These systems are efficient in detaching apples. 
	However, they could not avoid bruising caused by apple-to-apple, apple-to-tree, and apple-to-container collisions, and hence they have not been adopted by the apple industry \cite{he2017}.
	On the other hand, fruit-by-fruit selective harvesting systems are developed with the aid of mechatronics and robotic technologies, which generally consist of a vision-based perception component, a manipulator, and an end-effector. The manipulator of such systems picks apples sequentially, and thus the fruit damage can be reduced substantially. However, the fruit-by-fruit harvesting system requires multi-disciplinary advances to enable various synergistic functionalities, including perception, manipulation, and control. Specifically, the perception module typically exploits sensors like cameras and lidars mounted on the robot to detect and localize the apples \cite{gongal2015,plebe2001,sa2016}. With the target positions provided by the perception system, the control system directs the manipulator to approach the fruit. A specialized end-effector (e.g., gripper or vacuum tube) then detaches the apple from the tree and drop it to a receiving device or container \cite{lu2018system}. As fruit-by-fruit robotic picking is more appealing to the apple industry, it will be the focus of this paper.
	
	Over the past two decades, several fruit-by-fruit robotic apple harvesting systems have been developed \cite{baeten2008,de2011,Davidson2016IROS,silwal2017}.
	For example, a 7 degree-of-freedom (DOF) industrial manipulator and a silicone funnel shaped gripper with an internal camera are developed in \cite{baeten2008}. The system scans the orchard canopy from 40 look-out positions, and for each position the ripe apples are detected and picked one-by-one in a looped task. In \cite{silwal2017}, an apple harvesting robot is developed with a global camera, a 7 DOF manipulator, and a finger-based end-effector. Instead of attaching the camera to the end-effector as in \cite{baeten2008} and \cite{de2011}, the global camera, independent of other parts of the harvesting system, is employed to provide a larger field of view. 
	For both robotic systems reported in \cite{baeten2008} and \cite{silwal2017}, a 7 DOF manipulator is required to approach the fruit.
	While the 7 DOF manipulator can provide high maneuverability, it is overly complicated and extravagant for practical use. 
	
	During the apple picking process, the manipulator needs to approach fruits located at various positions within the workspace, and it thus requires a robust and accurate motion control scheme. 
	Several advances have been made in manipulator control for robotic harvesting. 
	For example, a two-step control method is developed in \cite{baeten2008}, where the manipulator is initially adjusted such that the camera's optical axis points straight to the apple and it is then controlled to reach out to the apple along the optical axis. This two-step method can be used in unstructured orchard environments, but it leads to discontinuous manipulation motion and low harvesting efficiency. 
	Another control scheme is developed in \cite{silwal2017}, which regulates the end-effector along a horizontal path or $45^{\circ}$ inclined path to reach the fruits. This scheme is only effective for V-trellis orchard architectures and would not be suitable for other modern structured orchards. 
	
	Despite the aforementioned efforts, there are still no commercially available robotic harvesting systems for tree fruits because the developed systems are still unsatisfactory in performance, too complicated or expensive to be economically viable, and unreliable or inefficient for working in the real orchard environment \cite{lu2017innovative,gongal2015}. To lay a foundation for automated apple harvesting, this paper presents the development of a new robotic apple harvesting prototype. 
	Specifically, an RGB-D camera is exploited as the primary sensor, and a deep learning-based perception module is developed for apple detection and localization. A 3 DOF manipulator and a vacuum-based end-effector are designed to approach and detach the apple, respectively. Different from previous studies that rely on high DOF industrial manipulators, the developed 3 DOF manipulator has simple and compact structure while providing high picking efficiency. Furthermore, a motion control scheme is designed to ensure the manipulator can approach apples with satisfactory accuracy. 
	Experiments are conducted to illustrate the performance of the integrated system.
	
	The main contributions of this paper include the following.
	Firstly,  we present the synergistic development of a robotic apple harvesting prototype that is simple in design, fast in actuation, and efficient in fruit picking.
	Secondly, the nonlinear control strategy is designed by fully exploiting the mechanical structure of the manipulator, which can avoid discontinuous manipulation motion and accomplish more agile apple approaching compared to \cite{baeten2008,silwal2017}. Last but not least, the experimental studies validate the design concept and underpin future research.
	
	The remainder of this paper is organized as follows. 
	Section \ref{sec_mechDesign} presents the system design of the apple harvesting robot whereas Section \ref{sec_motionControl} details the motion control scheme. Experiment results are provided in Section \ref{sec_perfEva}. Finally, conclusions are drawn in Section \ref{sec_conclusion}.

	\section{System Design} \label{sec_mechDesign}	
\begin{figure}[!h]
		\centering
		\includegraphics[width=6.5cm]{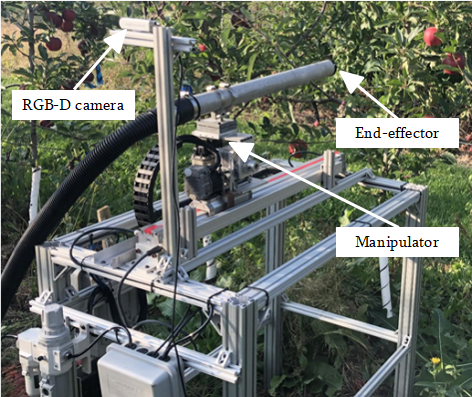}
		\caption{The developed robotic apple harvesting prototype.}\label{fig_appleRob}
	\end{figure}

	Our developed harvesting robot prototype is illustrated in Fig. \ref{fig_appleRob}. The hardware consists primarily of three modules: an Intel RealSense RGB-D camera, a 3 DOF manipulator, and a vacuum-based end-effector. Auxiliary units (e.g., power supply, vacuum pump) are placed on the rear side of the system (near the left lower corner in the image). 
	Those units are connected to a laptop (Intel i5‐6700 CPU and 16 GB RAM). 
	The robot operating system (ROS) is utilized to fully integrate the entire system and facilitate the communication and control of the modules. Below is a detailed description of each module. 
	
	\subsection{Visual Perception} \label{sec_visualSensing}
	
	The first and foremost task of automated apple harvesting is the detection and localization of fruits on the tree. Apple detection is to segment apples from the background (i.e., foliage, branches and trunks), while fruit localization refers to calculating the three-dimensional spatial position of the detected apples relative to the camera frame. In our system, an Intel RealSense D435i RGB-D camera is used to capture the environmental information, because of its compactness, low cost, and accuracy ($1280\times 720$ active stereo depth resolution \cite{RealSense}).
	\begin{figure}[!h]
		\centering
		\includegraphics[width=7cm]{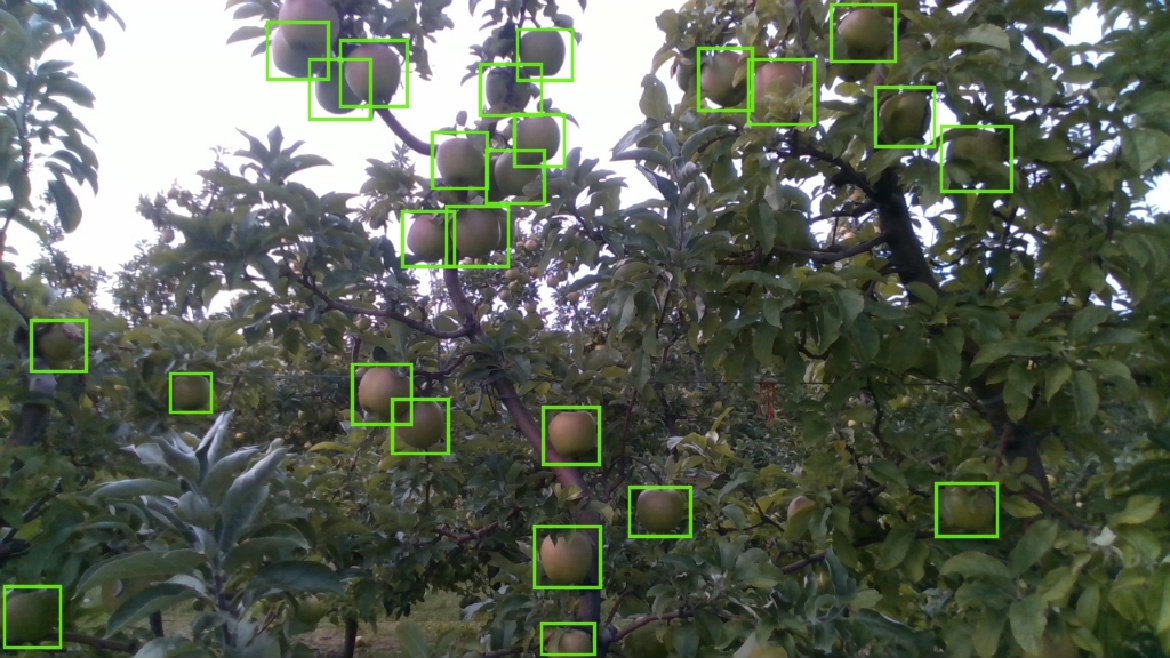}
		\caption{An example of using the Mask R-CNN based algorithm for detecting 'Gala' apples, where green bounding boxes represent identified apples.}\label{fig_appleDet}
	\end{figure}
	
	
	Apple detection and localization in the complex orchard environment is a challenging task due to partial occlusions by foliage and branches, varying lighting conditions, and color variations in different varieties and ripeness.
	To address the above challenges and achieve robust and accurate apple detection, a deep learning method based on Mask R-CNN \cite{he2017mask} is utilized. Mask R-CNN is the state-of-the-art deep neural network-based object detection algorithm that has found great successes in various applications, including vehicle detection \cite{barea2019}, nuclei segmentation \cite{VuolaISBI2019}, and fruit detection \cite{ganesh2019deep}. It exploits a mask branch network to enhance the end-to-end classification and segmentation capacities. 
	To train the network, we collected a comprehensive orchard image dataset for two apple varieties (i.e., 'Gala' and 'Blondee') under sunny and cloudy weather conditions for different time periods of the day (9 am, noon, and 3 pm) from a commercial orchard in Sparta, Michigan, USA during the 2019 harvest season.
	A total of 1,243 images were collected, among which 933 images were used for training the Mask R-CNN while the remaining for validation. The detection algorithm based on Mask R-CNN achieved a fruit identification accuracy of 92.7\% in the test dataset. A detailed report on the implementation of the mask R-CNN algorithm for apple detection and localization is given in \cite{Chu2020PRL}.
	Fig. \ref{fig_appleDet} shows an example of the detection results where green boxes show the identified apples.

	With the detected apples in bounding boxes, apple locations are computed by incorporating the depth information in the Intel RealSense RGB-D camera. 
	Specifically, for each detected apple, the disparity map is leveraged to generate a range matrix in the corresponding bounding box.
	The mean value of the range matrix is then calculated as the apple's depth range. Combining the depth range with the center of the bounding box pixels, the Cartesian position of the apple can be determined via back-projection \cite{Hartley2003}. This iterative process is run for each apple area to obtain positions for all detected apples in the image.
	
	
	\subsection{Manipulator Design} \label{subsec_manipulator}
\begin{figure}[!h]
		\centering
		\includegraphics[width=6.5cm]{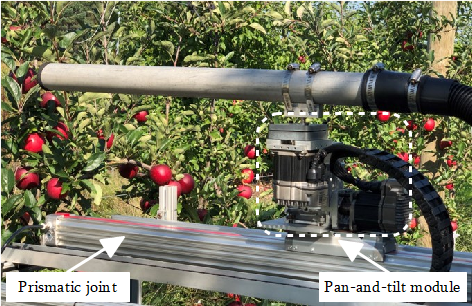}
		\caption{Proposed 3 DOF manipulator with two revolute joints and one prismatic joint.}\label{fig_fullManipulator}
	\end{figure}

	With the target apple locations provided by the perception system discussed above, a 3 DOF manipulator is then designed and assembled to efficiently reach the target locations. 
	As presented in Fig. \ref{fig_fullManipulator}, the manipulator consists of two revolute joints and one prismatic joint. The two revolute joints create a pan-and-tilt mechanism and are affixed to the prismatic base.
	This design provides a simple and compact mechanical structure, which not only offers sufficient DOF for primary pick and place tasks but also facilitates highly efficient motion control. 
	
	
	
\begin{figure}[!h]
		\centering
		\includegraphics[width=6.5cm]{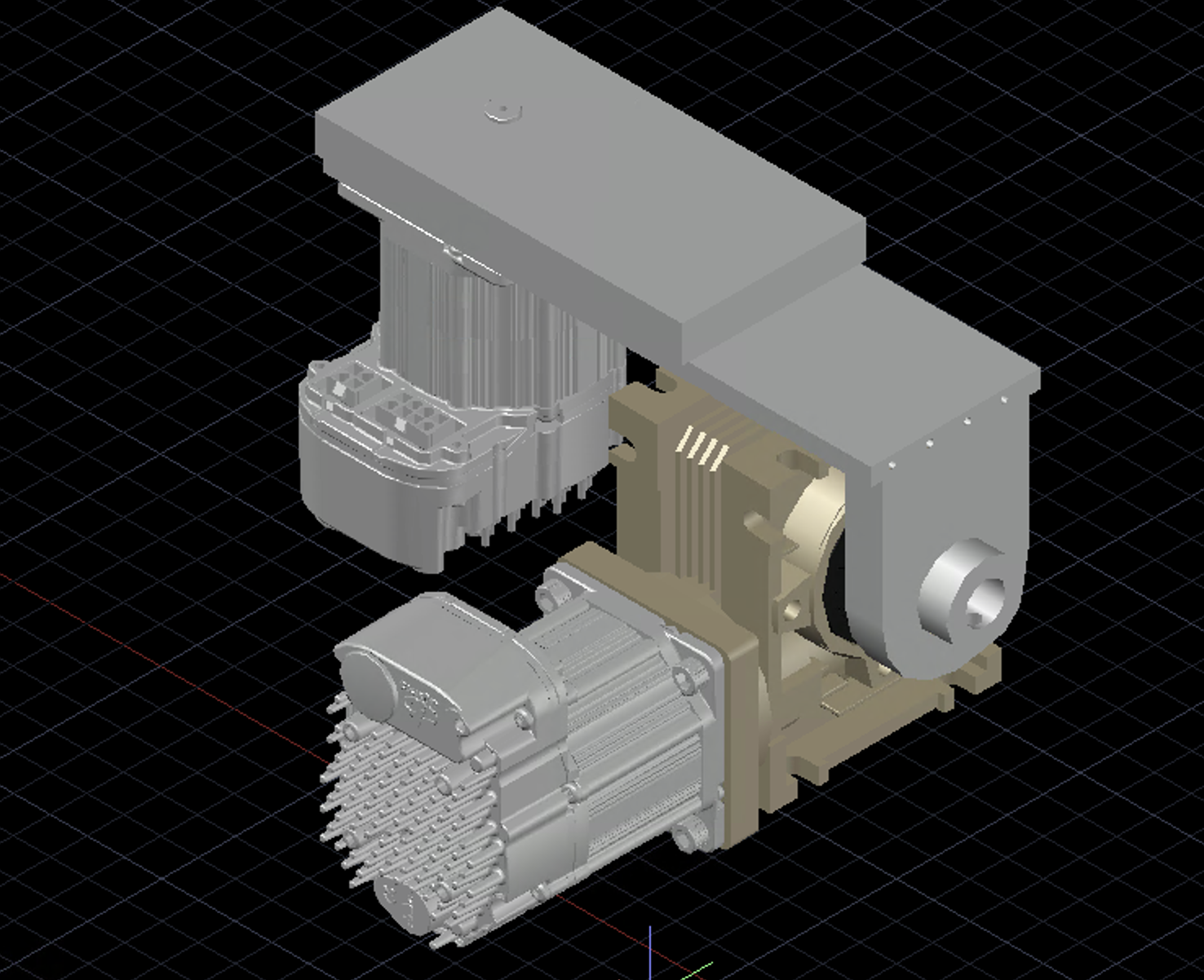}
		\caption{CAD model of the pan-and-tilt module constructed with two revolute joints.}\label{fig_panTilt}
	\end{figure}

	The pan-and-tilt mechanism is versatile and thus has been widely used in robotic systems \cite{stanfordArmBook,LeeTMECH2013}. As shown in Fig. \ref{fig_panTilt}, the pan-and-tilt module in our system contains two revolute joints that are driven by NEMA 23 Teknic ClearPath Servos motors operating at a maximum velocity of 4,000 RPM and peak torque of 2 N$\cdot$m. The tilt (vertical) joint is driven through a 90 degree worm gearbox with an 80:1 ratio and 10.17 N$\cdot$m holding torque, while the pan (horizontal) movement relies on a parallel shaft gearbox with a 45:1 ratio and 10.02 N$\cdot$m holding torque. The parallel shaft gearbox is a Molon gearbox modified to accommodate the motor's large drive shaft. 
	There two revolute joints are linked using a $L$-shaped aluminum plate, so that the axes of rotation of the two gearboxes are perpendicular to each other.
	
	The velocity of the revolute joints (i.e., servo motors) can be adjusted via variable frequency pulses ranging from 0 to 500 kHz. The pulse signals are generated by an Arduino Uno micro-controller. Based on the serial node provided by the ROS environment, communications between the Arduino interfaces and the servo motors are established. Furthermore, to achieve closed-loop control, the position feedback of the revolute joints needs to be measured. The position information cannot be accessed through the peripheral I/O ports of the servo motors, and thus it is necessary to introduce an adscititious sensing scheme. By default, the motor's user-settable counts per revolution gives an exact representation of the distance that the shaft travels per pulse. Hence, counting the pulses can deduce the position information of the revolute joints. Based on this observation, a Teensy 3.6 micro-controller is used as a counter of the pulse signals, and the real-time position information of the revolute joints is calculated with the counting results. The Teensy 3.6 micro-controller is running at the clock rate of 256 MHz, which can provide an accurate signals counting.
	
	
	
	As shown in Fig. \ref{fig_fullManipulator}, a prismatic joint is added as the base of the pan-and-tilt module to extend the depth of the manipulator's workspace. Specifically, the prismatic joint is a pneumatically actuated Lintra rodless air cylinder with a stroke length of 0.61 m and a slide carriage. The pneumatic system is driven by a 30-gallon air compressor, which enables the slide carriage to travel the entire stroke length in less than one second. High speed is the main advantage and reason for choosing pneumatic actuation over a screw based linear stage or a rack-and-pinion system. Moreover, the Enfield Technologies S2 Valve Positioning System allows for easy control of the prismatic joint through a standard voltage scheme. The carriage position along the stroke length can be read from a Balluff BTL6 MicroPulse transducer, and the control signals are generated with Arduino and ROS interfaces.
	
	
	Finally, a hollow aluminum link is installed on the pan-and-tilt module to make sure that the end-effector can reach the apple locations. The length and diameter of the link are 0.71 m and 0.04 m, respectively. This link also acts as a vacuum tube for grasping apple fruits in the harvesting process as referenced in \cite{Lu2020patent}. 
	
	
	
	\subsection{End-effector Design}
	Common issues in fruit harvesting end-effector designs include failure to isolate clustered fruit \cite{KONDO201020}, insufficient gripping strength \cite{bamotra}, low harvesting efficiency from high cycle times \cite{Davidson}, and damage to the fruit, canopy structures, or the end-effector itself due to bulky mechanical components \cite{Hemming}. Thus, end-effector design is a significant challenging task for researchers, and a wide range of design concepts have been studied with varying degrees of effectiveness and efficiency. In our system, a vacuum-based end-effector is utilized. It has been shown that the vacuum-based end-effector is effective in grasping and detaching tree fruits while minimizing bruising \cite{Gerber1985patent}. Additionally, when the manipulator does not approach the apple accurately, the vacuum-based end-effector can tolerate the approaching error since it can attract the fruit within a certain distance when sufficient vacuum flow is provided. This is important for in-field applications where unpredictable environmental factors (e.g., instantaneous movement of fruits due to disturbances from winds and the traveling robot platform, uneven orchard terrain, etc.) may adversely affect the system performance (i.e., accuracy in fruit localization and robot control and movement). 
	Selection of an appropriate diameter of the end-effector is critical, so that it can achieve a proper flow rate and vacuum pressure that is needed to grip and detach fruits of different size, while maintaining the agility of navigating in and out of the tree canopies. Through preliminary studies, the end-effector diameter of 0.04 m is determined to be adequate for the current robotic system.
	
	The rear end of the end-effector tube of the robot is connected to a Craftsman electric powered wet/dry vacuum via a flexible and expandable tube. During fruit picking, the vacuum machine operates in continuous mode, which can generate a peak horsepower of 5.5 HP.

	\section{Motion Control} \label{sec_motionControl}
	
	For an apple harvesting system, the manipulator needs to approach the apples located at different positions with high accuracy and flexibility. To achieve this goal, a motion control strategy is presented in this section by fully exploiting the mechanical structure of the developed 3 DOF manipulator.
	
	\subsection{Kinematic Model}
	\begin{figure}[!h]
		\centering
		\includegraphics[width=8.5cm]{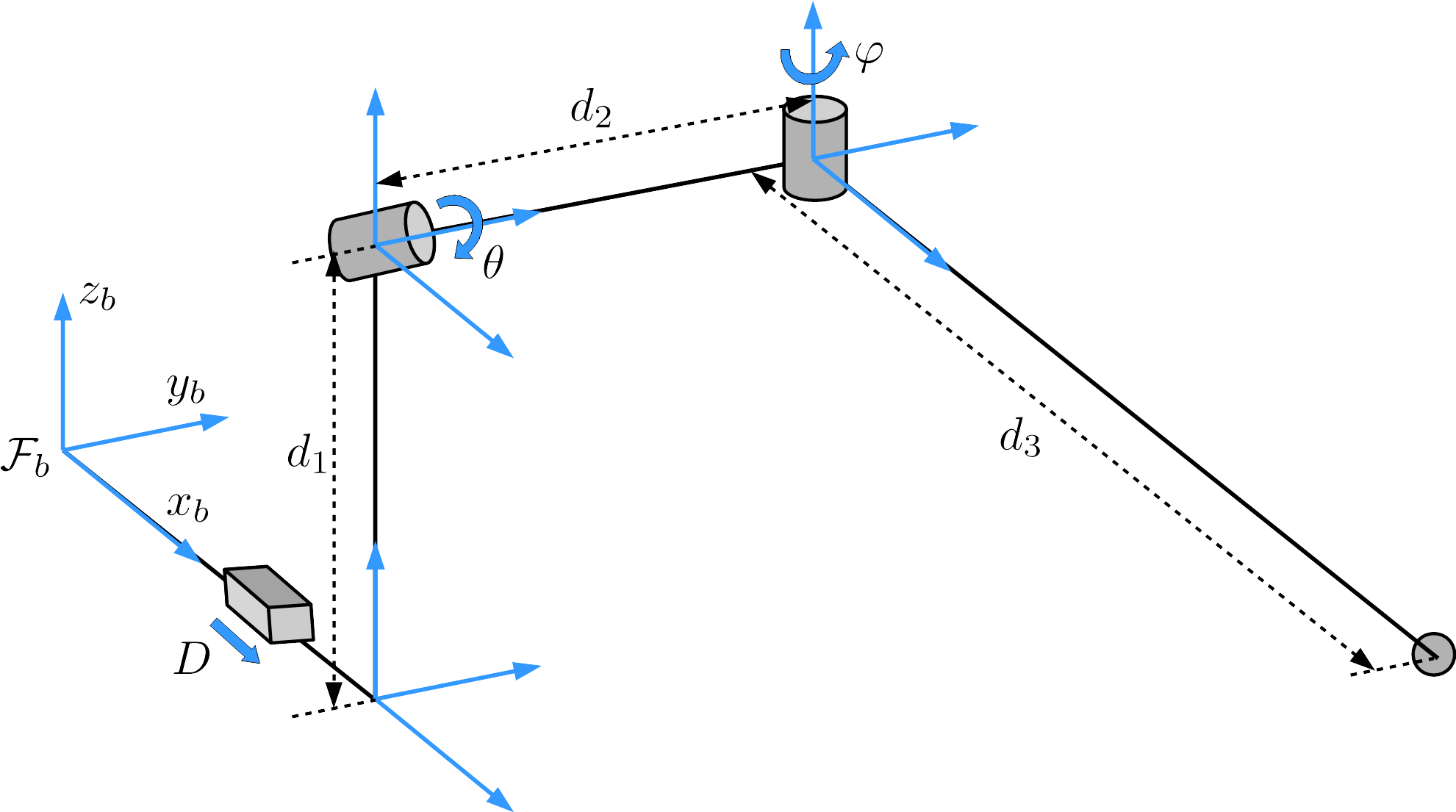}
		\caption{Kinematical description of the 3 DOF manipulator.}\label{fig_kineModel}
	\end{figure}
    
	The kinematical description of the 3 DOF manipulator is shown in Fig. \ref{fig_kineModel}. Let $\begin{bmatrix}
	x, y, z
	\end{bmatrix}^{T} \in 
	\mathbb{R}^{3}$ be the position of the end-effector expressed in terms of the base frame $\mathcal{F}_{b}$ of the manipulator. According to the kinematical diagram shown in Fig. \ref{fig_kineModel} and the Denavit–Hartenberg convention \cite{siciliano2010robotics}, the following forward kinematics function can be obtained:
	\begin{equation} \label{eq_xyz}
	\begin{aligned}
	x &= d_{3}\cos(\theta) \cos(\varphi) + D,
	\\
	y &= d_{3}\sin(\varphi) + d_{2},
	\\
	z &= -d_{3}\sin(\theta) \cos(\varphi) + d_{1},
	\end{aligned}
	\end{equation}
	where $d_{1}$, $d_{2}$, $d_{3} \in \mathbb{R}$ are the link length, and $\begin{bmatrix}
	\varphi, \theta, D
	\end{bmatrix}^{T} \in \mathbb{R}^{3}$ are the joint variables. The values of link length and joint variables are listed in Table \ref{table model parameter}. It is clear that \eqref{eq_xyz} characterizes the position of the end-effector as a function of the joint parameters $\begin{bmatrix}
	\varphi, \theta, D
	\end{bmatrix}^{T}$. From \eqref{eq_xyz} and the facts that $\varphi$, $\theta \in \left(-25^{\circ}, 25^{\circ} \right)$, it turns out that
	\begin{equation} \label{eq_inverKin}
	\begin{aligned}
	\varphi &= \arcsin (\frac{y-d_{2}}{d_{3}}),
	\\
	\theta &= \arcsin (\frac{d_{1}-z}{d_{3}\cos(\varphi)}),
	\\
	D &= x-d_{3}\cos(\theta)\cos(\varphi),
	\end{aligned}
	\end{equation}
	which characterizes the inverse kinematics to calculate the joint parameters $\begin{bmatrix}
	\varphi, \theta, D
	\end{bmatrix}^{T}$ from the position of the end-effector $\begin{bmatrix}
	x, y, z
	\end{bmatrix}^{T}$. Generally, gradient-based optimization solvers \cite{wang1991,SugiharaTRO2011} can be used to compute the inverse kinematics of a manipulator. However, since the developed 3 DOF manipulator has a simple and exploitable structure, its inverse kinematics can be determined by the analytical expression in \eqref{eq_inverKin}, which can avoid iterative and complex optimization procedure that is more time consuming and can induce numerical errors.
	\begin{table}[!htbp]
		\caption{Model parameters of the 3 DOF manipulator}
		\label{table model parameter}
		\begin{center}
			\begin{tabular}{c c}
				\hline
				\hline
				Parameter & Value \\
				\hline
				Link $d_{1}$ & 0.0635 $m$ \\
				
				Link $d_{2}$ & 0.0889 $m$ \\
				
				Link $d_{3}$ & 0.6985 $m$ \\
				
				Revolute joint $\varphi$ & $\left(-25^{\circ}, 25^{\circ} \right)$ \\    
				
				Revolute joint $\theta$ & $\left(-25^{\circ}, 25^{\circ} \right)$ \\
				
				Prismatic joint $D$ & $\left(0 m, 0.61 m\right)$\\            
				\hline
				\hline
			\end{tabular}
		\end{center}
	\end{table}
	
	\subsection{Controller Development}
	As described in Section \ref{sec_visualSensing}, the proposed perception algorithm can provide the apple locations expressed in the camera frame. By using conventional calibration techniques \cite{Bougue2013cali}, the transformation matrix between the camera frame and the base frame of the manipulator can be determined. Then, the apple location under the manipulator coordinate frame can be calculated via the transformation matrix. Let $\begin{bmatrix}
	x_{d}, y_{d}, z_{d}
	\end{bmatrix}^{T} \in \mathbb{R}^{3}$ be the detected apple position. The control objective is to regulate the end-effector to approach the apple position $\begin{bmatrix}
	x_{d}, y_{d}, z_{d}
	\end{bmatrix}^{T}$ from the home position $\begin{bmatrix}
	x_{0}, y_{0}, z_{0}
	\end{bmatrix}^{T} \in \mathbb{R}^{3}$.
	Note that the revolute joint parameters $\varphi$, $\theta$ and prismatic joint parameter $D$ are driven by distinct dynamic mechanisms (motor-based and pneumatic), and hence different control schemes are designed for these two types of joints. Specifically, the revolute joints $\varphi$, $\theta$ are regulated with the velocity-based control method, while the position-based controller is used to adjust the prismatic joint $D$. The velocity-based control method can generate explicit speed command to smoothly adjust the revolute joints based on real-time position feedback, which is more accurate and robust than the position-based controller. However, the position feedback of the current pneumatic prismatic joint is not available, therefore, position-based control method is used for the prismatic joint.  
	
	The velocity-based control scheme for the revolute joints is presented next. Based on \eqref{eq_xyz}, it can be found that the end-effector position along the $y_{b}$-axis and $z_{b}$-axis is determined by $\varphi$ and $\theta$. Therefore, the revolute joints are driven to ensure that $\begin{bmatrix}
	y, z
	\end{bmatrix}^{T}$ converges to $\begin{bmatrix}
	y_{d}, z_{d}
	\end{bmatrix}^{T}$. To achieve this task, the quintic function \cite{siciliano2010robotics} is exploited to generate a reference trajectory $\begin{bmatrix}
	y_{r}, z_{r}
	\end{bmatrix}^{T} \in \mathbb{R}^{2}$ and then the aforementioned regulation problem will be transformed into a trajectory tracking problem. More precisely, the quintic function-based reference trajectory $\begin{bmatrix}
	y_{r}, z_{r}
	\end{bmatrix}^{T}$ has the following form:
	\begin{equation} \label{eq_yr_zr}
	\begin{aligned}
	y_{r}(t) &= a_{y0} + a_{y1}t + a_{y2}t^2 +a_{y3}t^3 + a_{y4}t^{4} + a_{y5}t^{5},
	\\
	z_{r}(t) &= a_{z0} + a_{z1}t + a_{z2}t^2 +a_{z3}t^3 + a_{z4}t^{4} + a_{z5}t^{5},
	\end{aligned}
	\end{equation}
	where $a_{yi}$, $a_{zi} \in \mathbb{R}$ $(i=0, 1, \cdots, 5)$ are coefficients of the quintic function and $t$ denotes time. 
	Given $\begin{bmatrix}
	y_{0}, z_{0}
	\end{bmatrix}^{T}$, $\begin{bmatrix}
	y_{d}, z_{d}
	\end{bmatrix}^{T}$, and a time domain $\left[0, t_{f}\right]$, the reference trajectory satisfies
	\begin{equation} \label{eq_yr_zr_cons}
	\begin{aligned}
	y_{r}(0) &= y_{0}, \quad y_{r}(t_{f}) = y_{d}, 
	\\
	\dot{y}_{r}(0) &= \ddot{y}_{r}(0) = \dot{y}_{r}(t_{f}) = \ddot{y}_{r}(t_{f}) = 0, 
	\\
	z_{r}(0) &= z_{0}, \quad z_{r}(t_{f}) = z_{d},
	\\
	\dot{z}_{r}(0) &= \ddot{z}_{r}(0) = \dot{z}_{r}(t_{f}) = \ddot{z}_{r}(t_{f}) = 0.
	\end{aligned}
	\end{equation}
	Based on \eqref{eq_yr_zr} and \eqref{eq_yr_zr_cons}, the coefficients $a_{yi}$, $a_{zi} (i=0, 1, \cdots, 5)$ can be calculated. Note that the constraints presented in \eqref{eq_yr_zr_cons} ensure that the initial and final positions of the reference trajectory will be $\begin{bmatrix}
	y_{0}, z_{0}
	\end{bmatrix}^{T}$ and $\begin{bmatrix}
	y_{d}, z_{d}
	\end{bmatrix}^{T}$, respectively. 
	Therefore, actuating the revolute joints to make $\begin{bmatrix}
	y, z
	\end{bmatrix}^{T}$ follow the reference trajectory $\begin{bmatrix}
	y_{r}, z_{r}
	\end{bmatrix}^{T}$ will lead to the convergence of $\begin{bmatrix}
	y, z
	\end{bmatrix}^{T}$ to $\begin{bmatrix}
	y_{d}, z_{d}
	\end{bmatrix}^{T}$. The introduction of the reference trajectory $\begin{bmatrix}
	y_{r}, z_{r}
	\end{bmatrix}^{T}$ also brings several additional advantages. First, the reference trajectory is continuously differentiable, which is conducive to ensuring that the end-effector approaches the desired position along a smooth path. Second, by adjusting the parameter $t_{f}$, the velocity profile of the reference trajectory can be modified, and thus the end-effector can reach the desired position within a specific time interval.  
	Based on \eqref{eq_xyz}, the time derivative of $\begin{bmatrix} y, z \end{bmatrix}^{T}$ can be calculated as:
	\begin{equation} \label{eq_dot_yz}
	\begin{aligned}
	\dot{y} &= d_{3}\cos(\varphi) \omega_{\varphi},
	\\
	\dot{z} &= -d_{3}\cos(\theta) \cos(\varphi) \omega_{\theta} + d_{3}\sin(\theta)\sin(\varphi) \omega_{\varphi},
	\end{aligned}
	\end{equation}
	where $\omega_{\varphi}$, $\omega_{\theta} \in \mathbb{R}$ are the angular velocity inputs of the revolute joints $\varphi$ and $\theta$, respectively. Furthermore, the error signals $\begin{bmatrix}
	e_{y}, e_{z}
	\end{bmatrix}^{T} \in \mathbb{R}^{2}$ are constructed as
	\begin{equation} \label{eq_ey_ez}
	\begin{aligned}
	e_{y} &= y-y_{r}, 
	\\
	e_{z} &= z-z_{r}.
	\end{aligned}
	\end{equation}
	Based on \eqref{eq_dot_yz}, \eqref{eq_ey_ez}, and by virtue of Lyapunov-based control techniques \cite{khalil2002nonlinear}, the velocity controller is designed as
	\begin{equation} \label{eq_omega}
	\begin{aligned}
	\omega_{\varphi} & = \frac{1}{d_{3}\cos(\varphi)} \left( -k_{1}e_{y} + \dot{y}_{r} \right),
	\\
	\omega_{\theta} & = \frac{1}{d_{3}\cos(\theta) \cos(\varphi)} \left( k_{2}e_{z} + d_{3}\sin(\theta)\sin(\varphi)\omega_{\varphi} - \dot{z}_{r} \right),
	\end{aligned}
	\end{equation}
	where $k_{1}$, $k_{2} \in \mathbb{R}^{+}$ are positive constant gains. 
	The velocity controller \eqref{eq_omega} can ensure that the end-effector position along the $y_{b}$-axis and $z_{b}$-axis tracks the reference trajectory $\begin{bmatrix}
	y_{r}, z_{r}
	\end{bmatrix}^{T}$ asymptotically, and the detailed stability analysis is given in Appendix A.
	
    We next present the position-based scheme for the prismatic joint control. As mentioned in Section \ref{subsec_manipulator}, the prismatic joint is driven by a pneumatic system. A voltage-based proportional-integral (PI) controller is utilized to regulate the prismatic joint parameter $D$. Specifically, given the desired position $\begin{bmatrix}
	x_{d}, y_{d}, z_{d}
	\end{bmatrix}^{T}$, the inverse kinematics \eqref{eq_inverKin} is used to calculate the corresponding desired joint parameters $\begin{bmatrix}
	\varphi_{d}, \theta_{d}, D_{d}
	\end{bmatrix}^{T}$. Then, based on the desired value $D_{d}$, the embedded PI controller can adjust the prismatic joint to ensure $D$ converges to $D_{d}$.  
	
	\section{Performance Evaluation} \label{sec_perfEva}
	
	In this section, comprehensive experiments are reported to demonstrate the performance of the developed robotic apple harvesting system. We first validate the motion control scheme developed in Section \ref{sec_motionControl} and then evaluate the integrated system in apple picking scenarios.
	
	\subsection{Motion Control Validation}
	Since different control schemes are employed for the revolute and prismatic joints, their performance is evaluated separately in the following. 
	
\begin{figure}[!h]
		\centering
		\includegraphics[width=6.5cm]{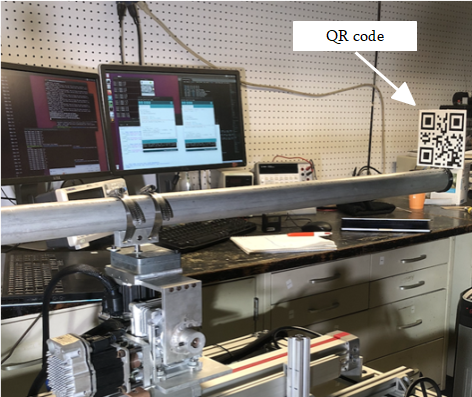}
		\caption{Experimental setup for control validation.}\label{fig_qrCode}
	\end{figure}

	The velocity controller \eqref{eq_omega} designed for the revolute joints $\varphi$ and $\theta$ is tested firstly. 
	We use open-loop velocity control and position control approaches as the benchmark to facilitate the performance evaluation. In particular, the open-loop velocity controller is given by
	\begin{equation} \label{eq_omega_ol}
	\begin{aligned}
	\omega_{\varphi} & = \frac{\dot{y}_{r}}{d_{3}\cos(\varphi)},
	\\
	\omega_{\theta} & = \frac{1}{d_{3}\cos(\theta) \cos(\varphi)} \left( d_{3}\sin(\theta)\sin(\varphi)\omega_{\varphi} - \dot{z}_{r} \right).
	\end{aligned}
	\end{equation}
	Comparing \eqref{eq_omega} with \eqref{eq_omega_ol}, it can be found that the developed controller exploits the feedback errors $\begin{bmatrix}
	e_{y}, e_{z}
	\end{bmatrix}^{T}$ to achieve closed-loop control, while the open-loop velocity controller does not include the feedback error terms. The position control method utilizes the positioning mode provided by the NEMA 23 Teknic ClearPath Servos to rotate the revolute joints. More specifically, given the desired position of the end-effector, the corresponding desired joint values $\varphi_{d}$ and $\theta_{d}$ can be calculated via the inverse kinematics \eqref{eq_inverKin}, and then the algorithm embedded in the positioning mode is called to regulate the servo motor towards the desired joint values. To conduct a thorough comparison, three cases are selected. Under each case, the manipulator is controlled by the aforementioned three methods from the same home position to a target position. The desired positions for these three cases are chosen as:
	\[
	\begin{aligned}
	\text{Case 1:} \begin{bmatrix}
	x_{d}, y_{d}, z_{d}
	\end{bmatrix}^{T} &= \begin{bmatrix}
	0.6876m, -0.0505m, 0.011m
	\end{bmatrix}^{T},
	\\
	\text{Case 2:} \begin{bmatrix}
	x_{d}, y_{d}, z_{d}
	\end{bmatrix}^{T} &= \begin{bmatrix}
	0.5874m, -0.1483m, 0.3695m
	\end{bmatrix}^{T},
	\\
	\text{Case 3:} \begin{bmatrix}
	x_{d}, y_{d}, z_{d}
	\end{bmatrix}^{T} &= \begin{bmatrix}
	0.6936m, 0.1938m, 0.01m
	\end{bmatrix}^{T}.
	\end{aligned}
	\] 
	According to \eqref{eq_inverKin}, the corresponding desired joint values can be calculated as follows:
	\[
	\begin{aligned}
	\text{Case 1:} \begin{bmatrix}
	\varphi_{d}, \theta_{d}, D_{d}
	\end{bmatrix}^{T} &= \begin{bmatrix}
	-11.5^{\circ}, 8.6^{\circ}, 0m
	\end{bmatrix}^{T},
	\\
	\text{Case 2:} \begin{bmatrix}
	\varphi_{d}, \theta_{d}, D_{d}
	\end{bmatrix}^{T} &= \begin{bmatrix}
	-19.8^{\circ}, -23.1^{\circ}, 0m
	\end{bmatrix}^{T},
	\\
	\text{Case 3:} \begin{bmatrix}
	\varphi_{d}, \theta_{d}, D_{d}
	\end{bmatrix}^{T} &= \begin{bmatrix}
	8.6^{\circ}, 8.6^{\circ}, 0m
	\end{bmatrix}^{T}.
	\end{aligned}
	\] 
	Note that the prismatic joint parameter $D_{d}$ is set as zero in these three cases, aimed at solely validating the control performance of the revolute joints. Moreover, to measure the control accuracy, as shown in Fig. \ref{fig_qrCode}, a QR code is attached to the end-effector. The QR code can be detected and localized by the RGB-D camera stably and precisely, and thus the final position of the end-effector can be determined via the QR code identification. Each control method is tested by running 5 times in each case, and the average distance errors between the final position of the end-effector and the given desired position are calculated to facilitate the evaluation. The corresponding results are shown in Table \ref{table distance error}. It can be seen that the proposed velocity control scheme actuates the revolute joints $\varphi$ and $\theta$ with higher accuracy for all three cases compared to the other two methods. 
	
	\begin{table}[!htbp]
		\caption{Comparison of the distance errors ($mm$) between the final position and desired position of the manipulator for different revolute joint control approaches}
		\label{table distance error}
		\begin{center}
			\begin{tabular}{c c c c}
				\hline
				\hline
				& \textbf{Case 1} & \textbf{Case 2} & \textbf{Case 3} \\
				\hline
				Open-loop velocity control
				& 7.1 & 14.7 & 10.8 \\
				
				Position control
				& 6.2 & 8.6 & 2.9 \\    
				
				Proposed velocity control
				& \textbf{4.8} & \textbf{8.0} & \textbf{1.9} \\            
				\hline
				\hline
			\end{tabular}
		\end{center}
	\end{table}
	
	The evaluation of the position control scheme for the prismatic joint $D$ is presented next. To separately test the prismatic joint, the revolute joints $\varphi$ and $\theta$ are set as zero while the prismatic joint $D$ is actuated to achieve the following desired values: 
	\[
	D_{d} = 0.1m, 0.2m, 0.3m.
	\]    
	The movement of the prismatic joint $D$ is measured with the aid of QR code, and the measurement results corresponding to the desired values above are $0.111m$, $0.208m$, and $0.311m$. Moreover, the position control scheme can regulate the prismatic joint $D$ to approach the given desired values within one second in different tests, which satisfies the speed requirement for practical applications. 
	
	\subsection{Apple Harvesting Validation}
\begin{figure}[!h]
		\centering
		\includegraphics[width=6.5cm]{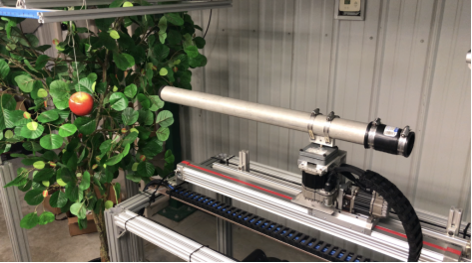}
		\caption{Experimental setup for apple harvesting validation.}\label{fig_applePick}
	\end{figure}

	To evaluate the performance of the integrated robotic apple harvesting system, picking tests are conducted in a laboratory setting with artificial apple trees as illustrated in Fig. \ref{fig_applePick}. An apple is hung on a movable aluminum frame. By adjusting the aluminum frame, the apple can be placed in arbitrary positions in the manipulator's workspace. The vision system is installed at the rear upper position of the manipulator to detect and localize the fruit. 	

	\begin{figure*}[!htbp]
		\centering
	    \subfigure[0s] {\label{0s}
	    	\includegraphics[width=0.15\textwidth]{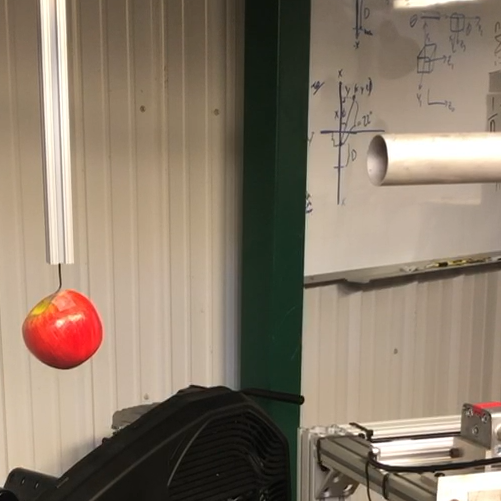}
    	}
	    \hspace{0.005 in}
	    \subfigure[0.4s] {\label{0.4s}
		    \includegraphics[width=0.15\textwidth]{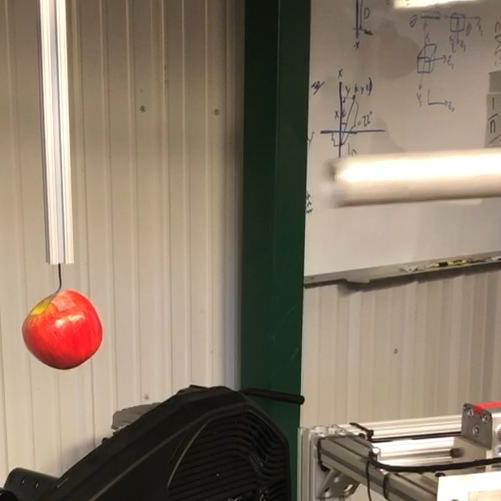}
	    }
	    \hspace{0.005 in}
	    \subfigure[0.8s] {\label{0.8s}
		    \includegraphics[width=0.15\textwidth]{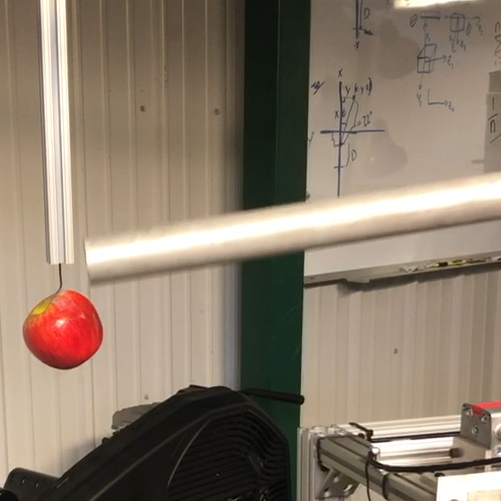}
	    }
	    \hspace{0.005 in}
	    \subfigure[1.2s] {\label{1.2s}
		    \includegraphics[width=0.15\textwidth]{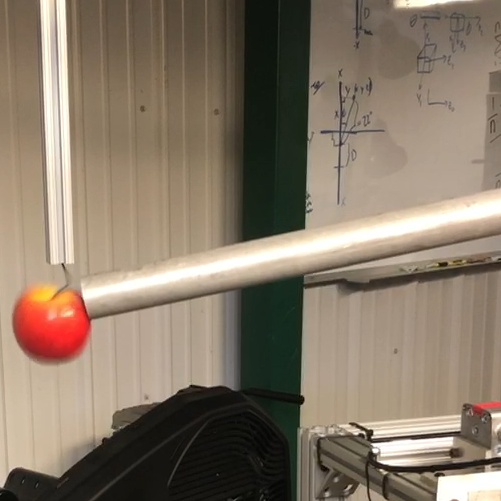}
	    }
	    \hspace{0.005 in}
	    \subfigure[1.6s] {\label{1.6s}
		    \includegraphics[width=0.15\textwidth]{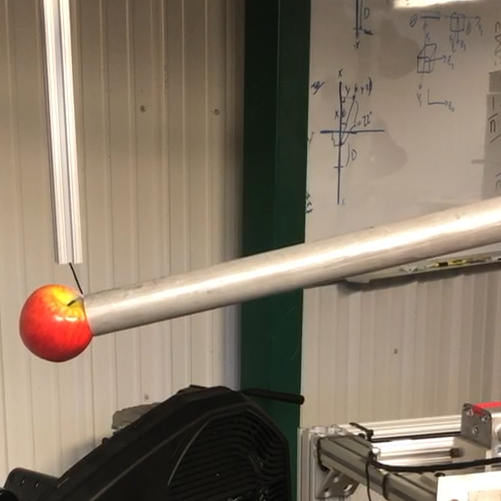}
	    }
	    \hspace{1 in}
	    \subfigure[2.4s] {\label{2.4s}
		    \includegraphics[width=0.15\textwidth]{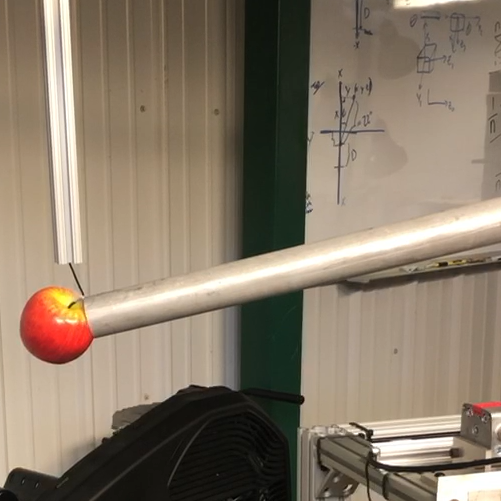}
	    }
	    \hspace{0.005 in}
	    \subfigure[2.8s] {\label{2.8s}
		    \includegraphics[width=0.15\textwidth]{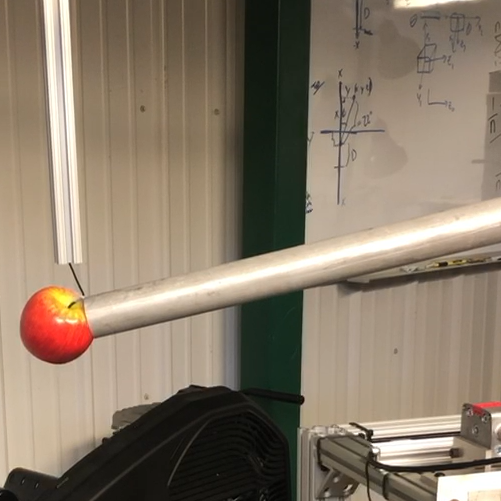}
	    }
	    \hspace{0.005 in}
	    \subfigure[3.2s] {\label{3.2s}
		    \includegraphics[width=0.15\textwidth]{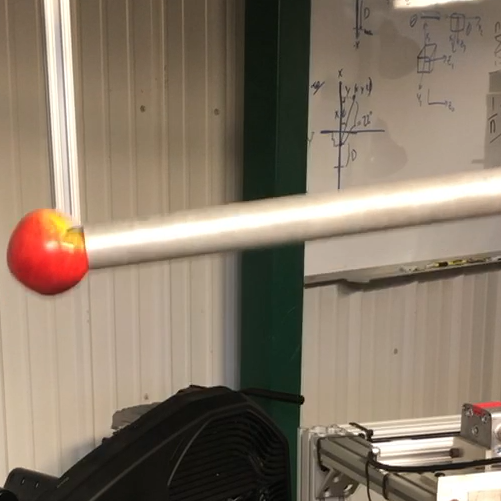}
	    }
	    \hspace{0.005 in}
	    \subfigure[3.6s] {\label{3.6s}
		    \includegraphics[width=0.15\textwidth]{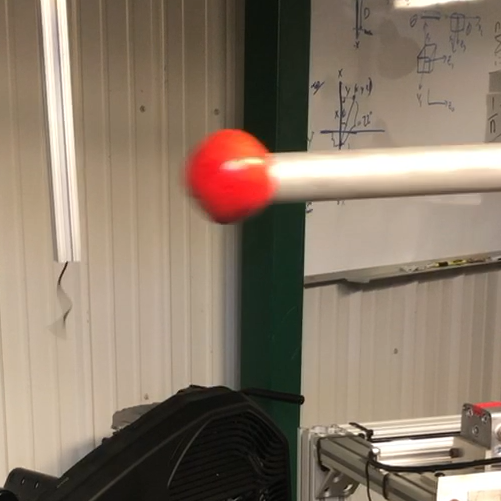}
	    }
	    \hspace{0.005 in}
	        \subfigure[4s] {\label{4s}
		    \includegraphics[width=0.15\textwidth]{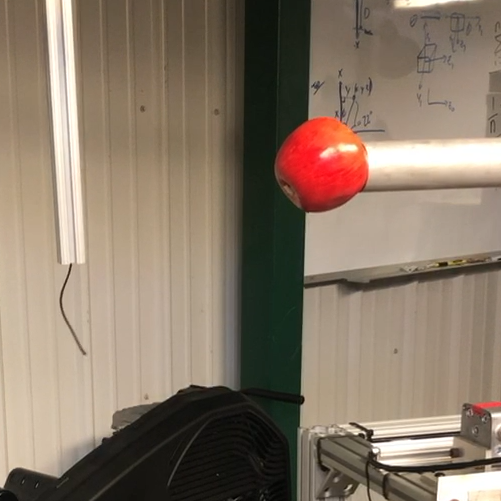}
	    }
	    \caption{Snapshots of a harvesting cycle at different time instants.}
	    \label{fig_pickCycle}
    \end{figure*}

\begin{figure}[!htbp]
	\centering
	\subfigure[] {\label{img_up}
		\includegraphics[width=0.14\textwidth]{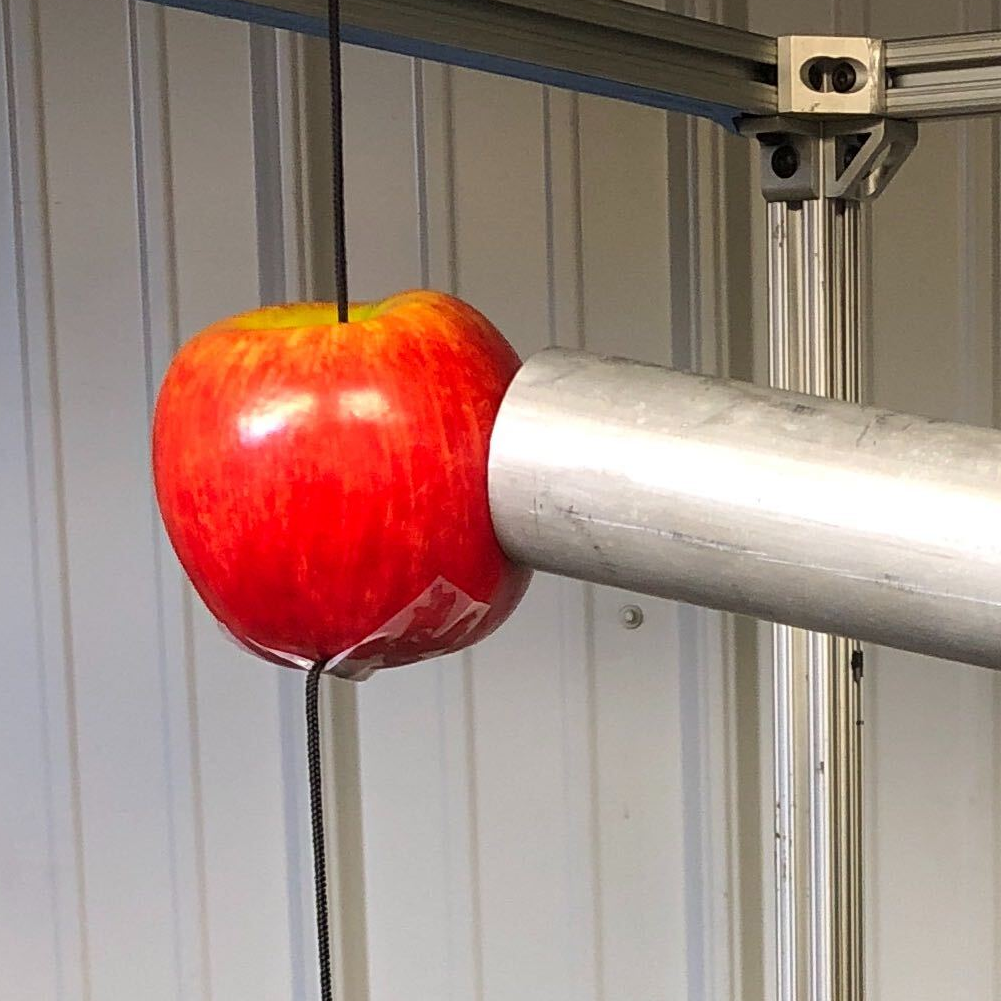}
	}
	\hspace{0.003 in}
	\subfigure[] {\label{img_mid}
		\includegraphics[width=0.14\textwidth]{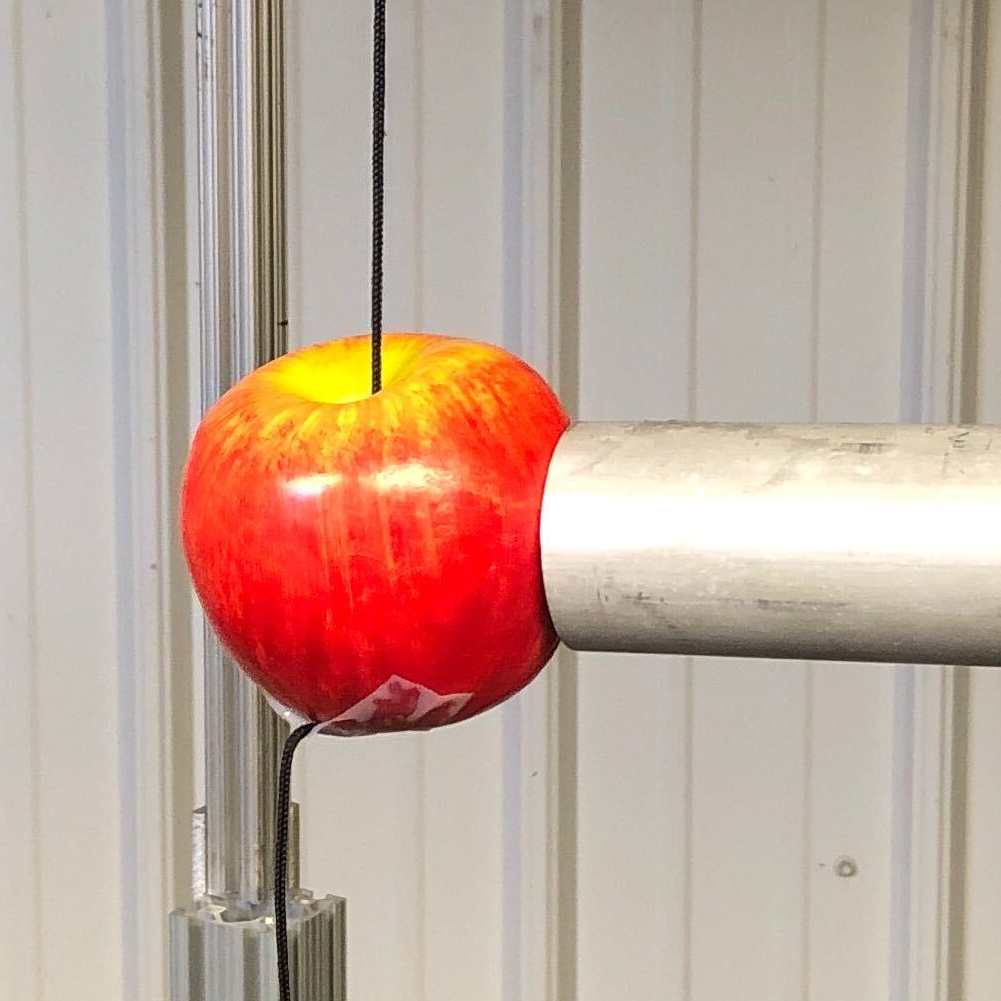}
	}
	\hspace{0.003 in}
	\subfigure[] {\label{img_down}
		\includegraphics[width=0.14\textwidth]{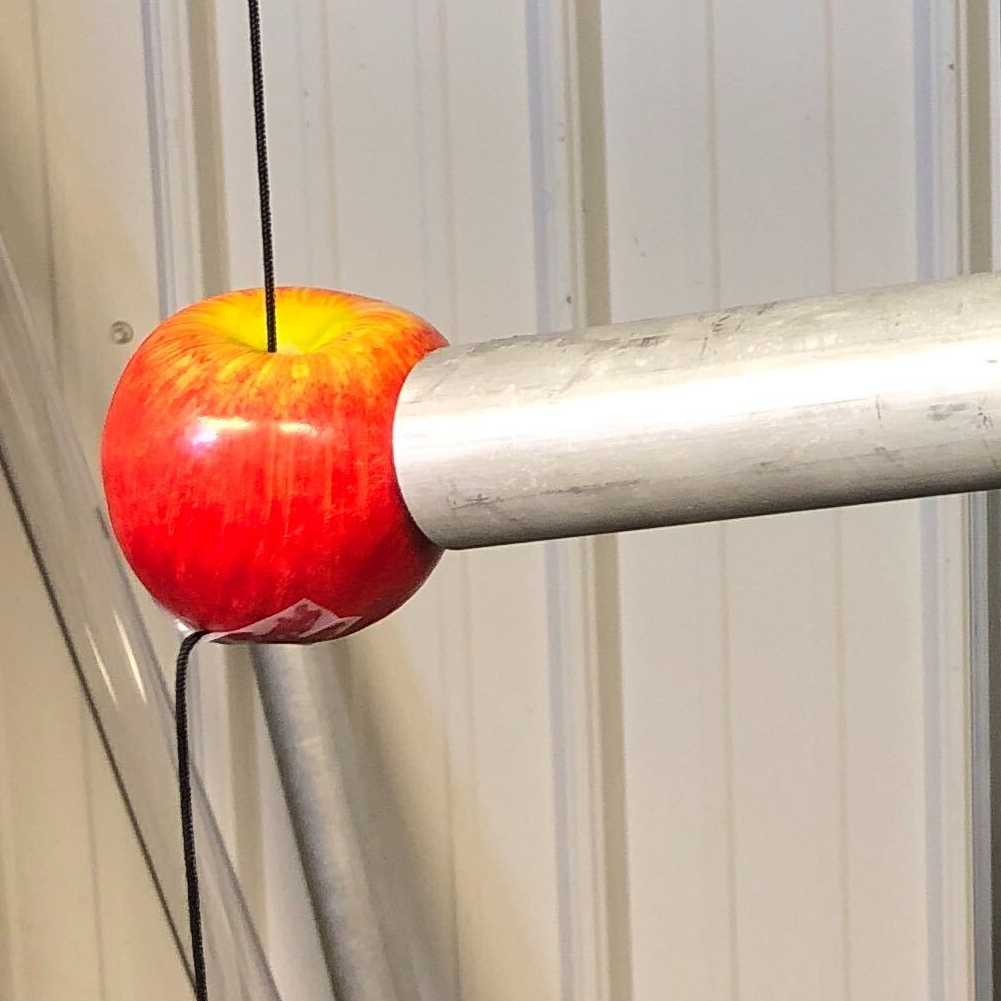}
	}
	\caption{Approaching performance of the manipulator with the apple located in different regions of the workspace.}
	\label{fig_approaching}
\end{figure}
	
	A total of 60 picking experiments are completed. In each picking experiment, the apple is randomly placed in the workspace. The vision system first identifies the apple and determines the apple location. Once the apple is localized, the manipulator is controlled to approach the fruit, and then the vacuum-based end-effector is actuated to detach the apple. Finally, the manipulator returns to the home position with the detached fruit. The snapshots of a complete picking cycle are presented in Fig. \ref{fig_pickCycle}, which shows that the developed robotic apple harvesting system can accomplish the primary picking functions. 
	Fig. \ref{fig_approaching} gives three samples of the harvesting tests, in which the manipulator follows different inclined angles to reach the apple located in diverse regions of the workspace.
	Different from \cite{silwal2017} which regulates the manipulator along fixed paths to approach a specific area, the developed control scheme can agilely adjust the manipulator to reach the apple arbitrarily placed in the workspace, which is a key capability for automated apple picking in structured/unstructured orchard environments. In all 60 harvesting tests, the manipulator approaches the apple accurately with the final approaching error being less than 2 cm, which is considered acceptable based on our prior laboratory and field tests. Within this error range, the vacuum-based end-effector can detach the fruit and firmly hold the fruit, as the manipulator is returning to the home position in all tests. Moreover, on average, 0.3 second is required to detect and localize all fruits in one image. The time for the manipulator to approach an individual apple is approximately 2.0 second, and fruit detaching using an open-loop command is set at 1.0 second.

    \subsection{Discussion on Future Work}
    While the developed prototype demonstrated promising performance, further work is needed to improve the system.
    
    
    It is necessary to introduce additional sensing to improve the picking efficiency and system robustness. The current command of fruit detaching operates in an open-loop scheme, which is unable to determine if or when the target fruit is detached from the tree. Installing a pressure sensor on the end-effector can provide feedback information about whether the fruit is held by the end-effector and has been separated from the tree after the predetermined movement, which is conducive to accomplishing closed-loop fruit detaching. 
    Besides, the perception system should be further extended to detect other objects, such as branches, to provide a more comprehensive environment perception. 
    
    Since many fruits are located deep in the canopy, a path planning algorithm needs to be developed and integrated with the motion controller to ensure that the manipulator approaches target apples without colliding or damage tree branches or other objects in its path to the target fruit.
    
    The current vacuum-based end-effector is made of aluminum tube covered with a thin layer of soft foam at its entrance. While this simple end-effector design is capable to generate sufficient vacuum pressure to detach about 80\% of apples from trees in our field tests in 2018 and 2019. Improvements to the end-effector design are needed, so as to achieve at least 95\% detaching rate. Different vacuum cup designs with different soft materials should be considered, so that the end effector can easily conform to the variable contours/sizes of apples to allow fast build-up of sufficient vacuum pressure to effectively detach fruits from trees.
    
	
	\section{Conclusion} \label{sec_conclusion}
	The mechatronic design and motion control of a robotic apple harvesting prototype was presented in this paper. This prototype integrated a vision-based perception system, 3 DOF manipulator, and vacuum-based end-effector to execute apple picking. A control scheme was designed to achieve accurate and agile manipulation motion. 
	Laboratory studies for 60 picking tests demonstrated that the manipulator reached the desired apple positions with the overall error being less than 2 cm, which is considered acceptable for the vacuum-based end-effector to detach fruits from trees. The developed prototype met the primary harvesting functionalities, thus laying a solid foundation for future advancements. 
	Future work will be focused on automated sensing of fruit holding and detaching process, optimal path planning for efficient picking of apples and minimizing potential damage to tree canopies by the manipulator, and better end-effector design for fast, firm holding and detaching of apples from trees.

	\appendices
	\section{Stability Analysis of the Velocity Controller}
	\begin{theorem}
		The velocity controller developed in \eqref{eq_omega} ensures that the end-effector position along the $y_{b}$-axis and $z_{b}$-axis, i.e., $\begin{bmatrix}
		y, z
		\end{bmatrix}^{T}$, converges to the reference trajectory $\begin{bmatrix}
		y_{d}, z_{d}
		\end{bmatrix}^{T}$ asymptotically. 
	\end{theorem}
	
	\begin{IEEEproof}
		To prove Theorem 1, a Lyapunov function $V \in \mathbb{R}$ is defined as 
		\begin{equation} \label{eq_V}
		V \triangleq \frac{1}{2}e_{y}^{2} + \frac{1}{2}e_{z}^{2},
		\end{equation}	
		where $e_{y}$ and $e_{z}$ are the error signals given in \eqref{eq_ey_ez}. Based on \eqref{eq_dot_yz} and \eqref{eq_ey_ez}, it can be obtained that
		\begin{equation} \label{eq_dot_ey_ez}
		\begin{aligned}
		\dot{e}_{y} &= d_{3}\cos(\varphi) \omega_{\varphi} - \dot{y}_{r},
		\\
		\dot{e}_{z} &= -d_{3}\cos(\theta) \cos(\varphi) \omega_{\theta} + d_{3}\sin(\theta)\sin(\varphi) \omega_{\varphi} - \dot{z}_{r}.
		\end{aligned}
		\end{equation}
		Taking the time derivative of \eqref{eq_V} and utilizing \eqref{eq_omega} and \eqref{eq_dot_ey_ez}, it can be further derived that
		\begin{equation} \label{eq_dot_V}
		\begin{aligned}
		\dot{V} &= e_{y}\dot{e}_{y} + e_{z}\dot{e}_{z}
		\\
		&= e_{y}\left( d_{3}\cos(\varphi) \omega_{\varphi} - \dot{y}_{r} \right) + e_{z} \left( -d_{3}\cos(\theta) \cos(\varphi) \omega_{\theta} \right. 
		\\
		&\quad \left. + d_{3}\sin(\theta)\sin(\varphi) \omega_{\varphi} - \dot{z}_{r} \right)
		\\
		&= -k_{1}e_{y}^{2} - k_{2}e_{z}^{2} \le 0.
		\end{aligned}
		\end{equation}
		According to \eqref{eq_V} and \eqref{eq_dot_V}, the Lyapunov's stability theorem \cite{khalil2002nonlinear} can be invoked to conclude that $e_{y}=0$ and $e_{z}=0$ are asymptotically stable, which indicates that $\begin{bmatrix}
		y, z
		\end{bmatrix}^{T}$ converges to $\begin{bmatrix}
		y_{r}, z_{r}
		\end{bmatrix}^{T}$ asymptotically.
	\end{IEEEproof}


	\ifCLASSOPTIONcaptionsoff
	\newpage
	\fi

	
	
	%
	\bibliographystyle{IEEEtran}
	\bibliography{IEEEabrv,reference}

\end{document}